\def\year{2020}\relax
\documentclass[letterpaper]{article} % DO NOT CHANGE THIS
\usepackage{aaai20}  % DO NOT CHANGE THIS
\usepackage{times}  % DO NOT CHANGE THIS
\usepackage{helvet} % DO NOT CHANGE THIS
\usepackage{courier}  % DO NOT CHANGE THIS
\usepackage[hyphens]{url}  % DO NOT CHANGE THIS
\usepackage{graphicx} % DO NOT CHANGE THIS
\usepackage{graphicx}  % DO NOT CHANGE THIS
\usepackage{algpseudocode}
\usepackage{algorithm}
\usepackage{algorithmicx}
\usepackage{algpseudocode}
\usepackage{amsmath}
\usepackage{amsfonts}
\usepackage{subcaption}
\usepackage{url}
\usepackage{array}
\usepackage{makecell}

\nocopyright
\title{Graduate Employment Prediction with Bias}
\author{
        Teng Guo,\textsuperscript{\rm 1}
        Feng Xia,\textsuperscript{\rm 1,2}
        Shihao Zhen,\textsuperscript{\rm 1}
        Xiaomei Bai,\textsuperscript{\rm 3}
        Dongyu Zhang,\textsuperscript{\rm 1}\thanks{Corresponding author: Dongyu Zhang (email: zhangdongyu@dlut.edu.cn)}
        Zitao Liu,\textsuperscript{\rm 4}
        Jiliang Tang\textsuperscript{\rm 5}\\
        \textsuperscript{\rm 1}Key Laboratory for Ubiquitous Network and Service Software of Liaoning Province,\\ Dalian University of Technology, Dalian 116620, China\\
        \textsuperscript{\rm 2}School of Science, Engineering and Information Technology, Federation University Australia, Ballarat, VIC 3353, Australia\\
        \textsuperscript{\rm 3}Computing Center, Anshan Normal University, Anshan 114007, China\\
        \textsuperscript{\rm 4}TAL AI Lab, TAL Education Group, Beijing 100080, China\\
        \textsuperscript{\rm 5}Department of Computer Science and Engineering, Michigan State University, East Lansing, MI 48824, USA\\
}
\begin{document}

\maketitle

\begin{abstract}
The failure of landing a job for college students could cause serious social consequences such as drunkenness and suicide. In addition to academic performance, unconscious biases can become one key obstacle for hunting jobs for graduating students. Thus, it is necessary to understand these unconscious biases so that we can help these students at an early stage with more personalized intervention. In this paper, we develop a framework, i.e., MAYA (\textbf{M}ulti-m\textbf{A}jor emplo\textbf{Y}ment st\textbf{A}tus) to predict students' employment status while considering biases. The framework consists of four major components. Firstly, we solve the heterogeneity of student courses by embedding academic performance into a unified space. Then, we apply a generative adversarial network (GAN) to overcome the class imbalance problem. Thirdly, we adopt Long Short-Term Memory (LSTM) with a novel dropout mechanism to comprehensively capture sequential information among semesters. Finally, we design a bias-based regularization to capture the job market biases. We conduct extensive experiments on a large-scale educational dataset and the results demonstrate the effectiveness of our prediction framework.
\end{abstract}

\section{Introduction}
Education, as the basic means of improving individual abilities, makes students competitive in recruitment. However, not every graduate can succeed in job hunting. The data from the statistical office of the European Union (EU) shows that the employment rate in the EU of 20-34 years old is 83.4\% for tertiary education and 65.8\% for upper secondary general education in 2018 \cite{eurostat}. The failure of job hunting could cause serious consequences like suicide \cite{drum2009new,westefeld2005perceptions}.
Therefore, detecting students with trouble in landing a job timely and providing personalized intervention and guidance at an early stage are greatly desired.

However, detecting these students faces tremendous challenges because recruitment might be impacted by various factors \cite{luo2018diagnosing}. Every recruiter aims to hire the best employees. However, in addition to academic performance \cite{kong2018analysis}, recruitment decisions can be affected by unconscious biases, such as gender \cite{clauset2015systematic}. These biases not only lead to imbalance in the hiring process, resulting in uniformity in the workplace instead of diversity, but also result in the inequality of employment \cite{giannakas2017hiring,ford2018gender,liang2018home}, especially for fresh graduates who have no work experience. Therefore, it is necessary to understand biases in recruitment, which can further be exploited for the prediction of graduates' employment. Nevertheless, previous related researches have been mainly based on questionnaires, which are time- and cost-consuming and hardly applicable to large-scale students.

Thanks to the advance of information technology, we are able to digitalize records of students in schools which product rich data about students.  It enables the data-driven development\cite{wu2013data,liu2018artificial} and provides us an opportunity to deepen our understandings on the employment of students. However, to achieve the goal, we face tremendous challenges. First, such data is much more complex than that based on questionnaires, thus advanced techniques are needed.   Second, the number of graduates that cannot land a job is much smaller compared to these who can successful obtain jobs, thus employment analysis and prediction are highly imbalanced. Third, there may exist biases in employment that can be varied by majors; while the majority of existing algorithms seldom consider possible biases.

In this paper, we are devoted to exploring the biases in different majors from demographics aspects, and predicting students with trouble in landing a job at an early stage. First, we analyze the employment biases for each major from 4 aspects including gender, nation, hometown, and enroll status. Second, based on possible employment biases, we propose a MAYA (\textbf{M}ulti-m\textbf{A}jor emplo\textbf{Y}ment st\textbf{A}tus) prediction framework, with four important components. In the first component, we solve the heterogeneity of students' courses through embedding academic performance into a space of unified dimension by autoencoder.
Then GAN (Generative Adversarial Networks) is applied to generate data of the minority class, to overcome the label imbalance problem.
Next, considering the sequential information between semesters, the Long Short-Term Memory (LSTM) with a novel dropout mechanism is utilized. Finally, we design a model to capture the employment biases of different majors. Our contributions can be summarized as follows:
\begin{itemize}
\item We provide a comprehensive and systematic analysis on employment biases.
\item We model the employment biases in different majors and incorporate them into our proposed prediction framework.
\item We conduct comprehensive experiments on a large-scale educational dataset and the extensive results demonstrate the effectiveness of our prediction framework.
\end{itemize}

This paper is organized as follows. In Section 2, related work is reviewed. The problem formulation is presented in Section 3. In Section 4, we analyze the employment biases by majors. In Section 5, the MAYA prediction framework is introduced in detail. In Section 6, we analyze the results of our experiment. We present the discussion and conclusion of our work in Section 7.

\section{Related Work}
\begin{figure*}
\centering
\includegraphics[width=0.51\columnwidth]{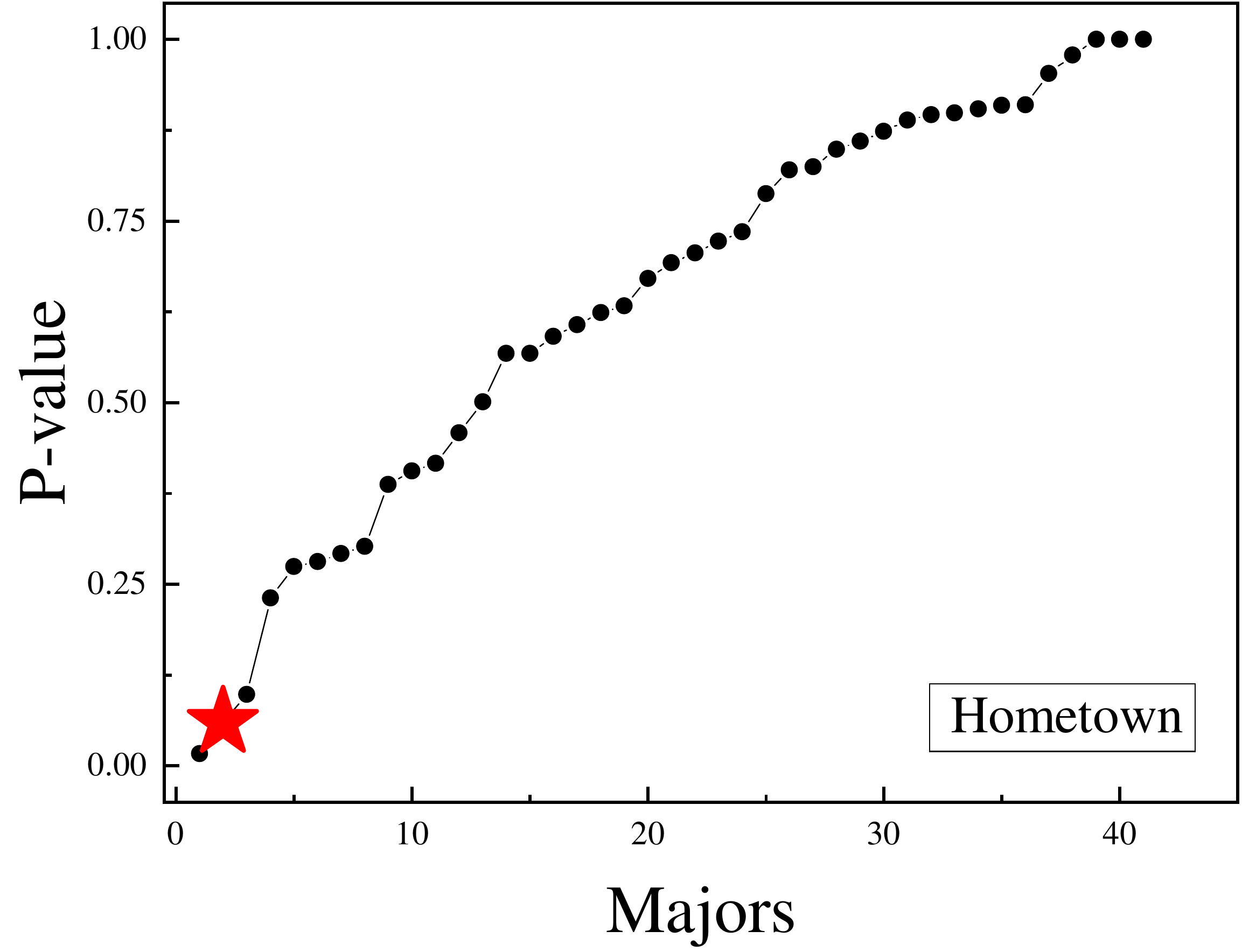}
\includegraphics[width=0.51\columnwidth]{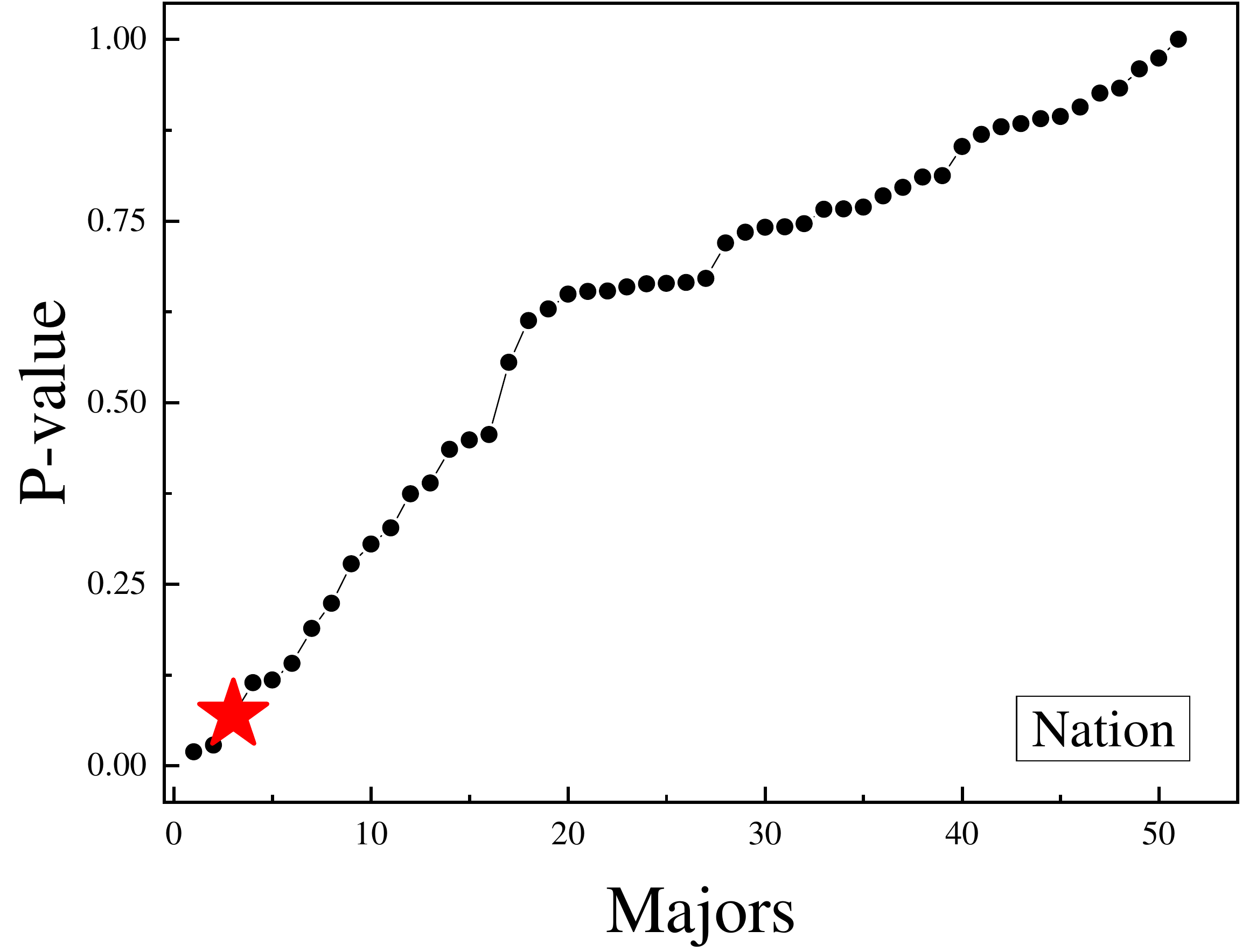}
\includegraphics[width=0.51\columnwidth]{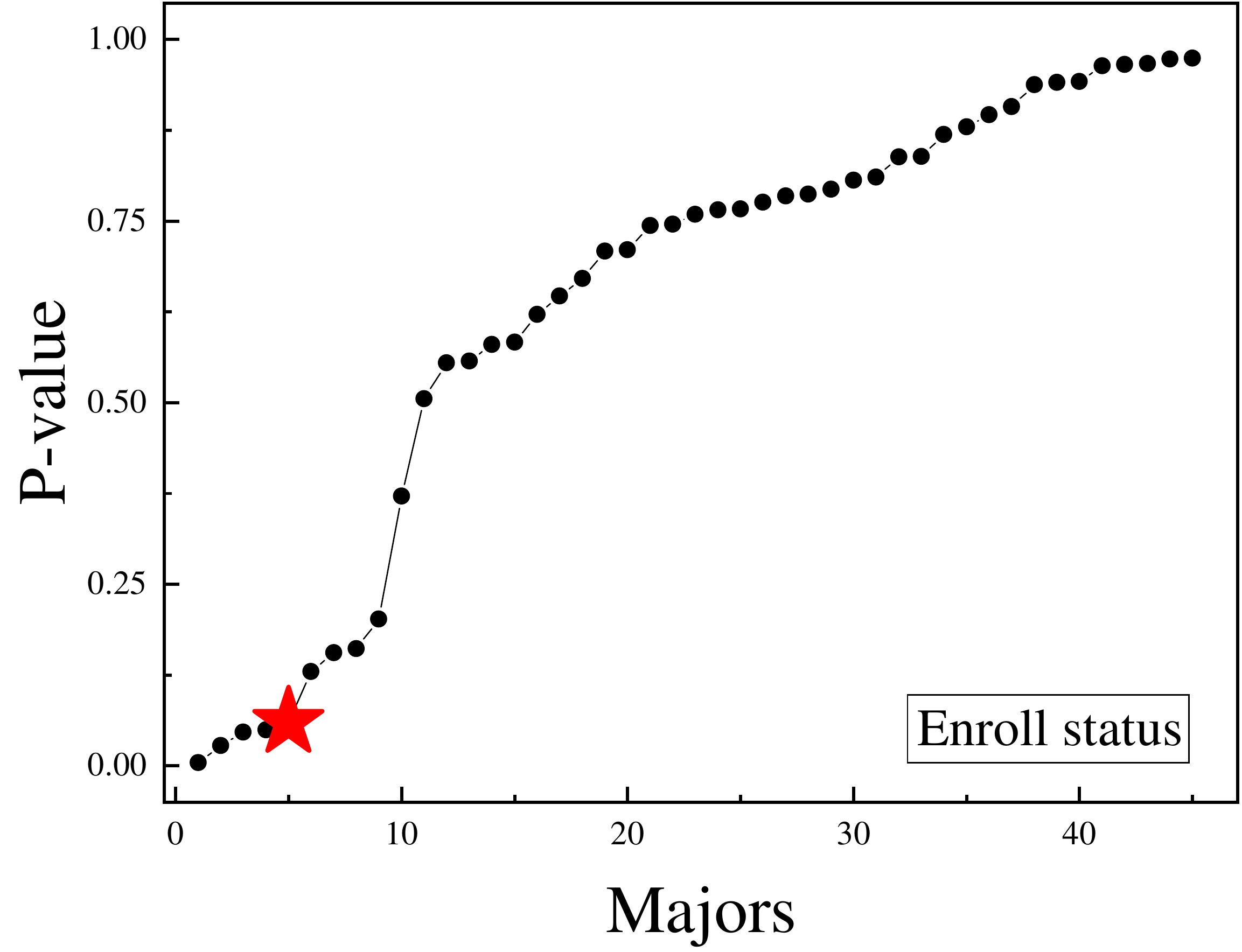}
\includegraphics[width=0.515\columnwidth]{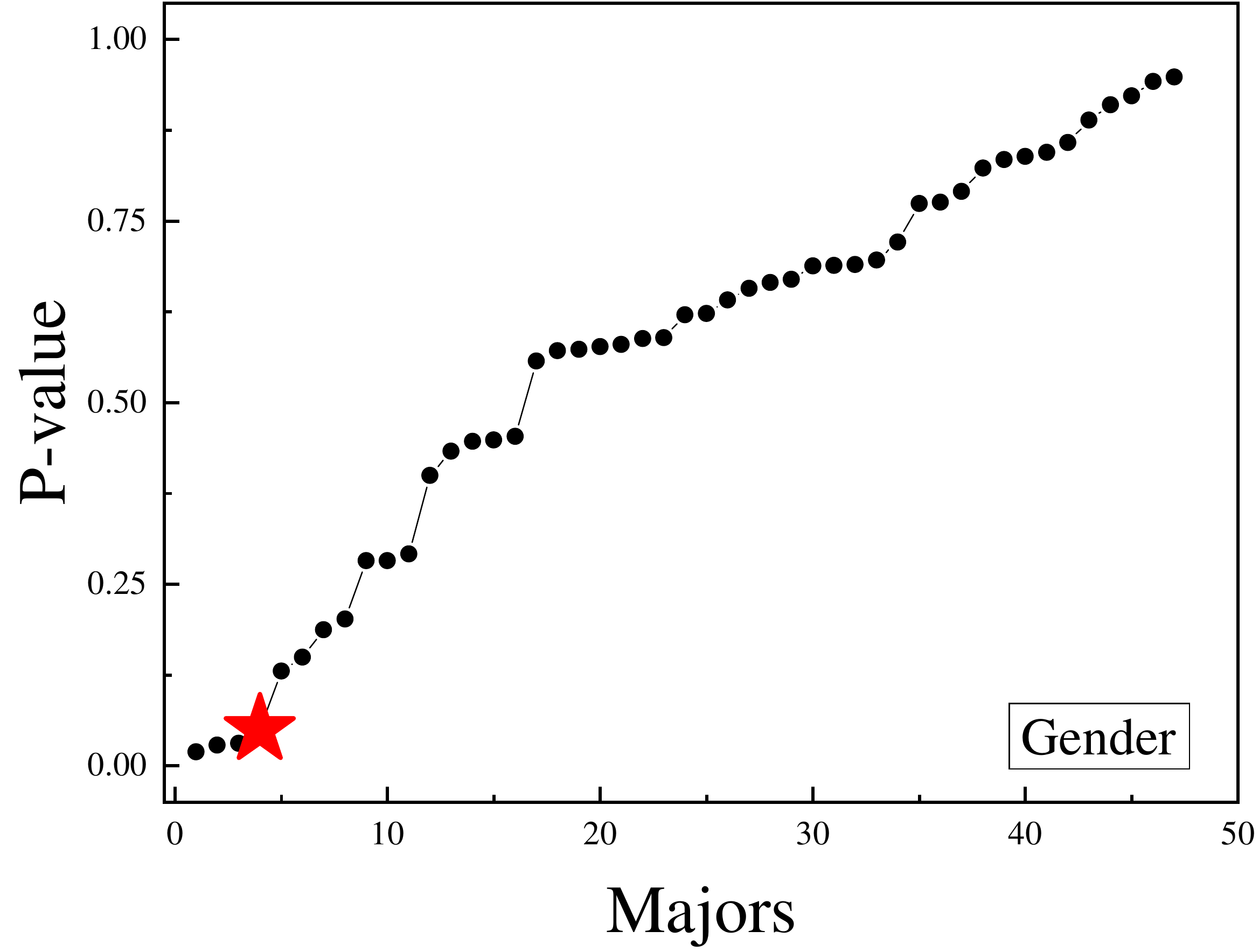}
\caption{The distribution of $p$-value with respective to majors on employment status. Subfigures denote the results of chi-square test in terms of hometown, nation, enroll status and gender, respectively. Each black dot represents the $p$-value of a certain major. When $p$-value is less than the threshold (e.g., 0.05 shown as red stars), the hypothesis is acceptable, that is, bias exists.}
	\label{bias-major}
\end{figure*}

\subsection{Employment of College Graduates}
Whether a college student can find a job after graduation attracts tons of attention in recent decades.
Liu et al \cite{liu2018application} develop a tool for career exploration based on the intuitiveness of node-link diagrams and the scalability of aggregation-based techniques to help the student understand the process of employment.
Kong et al \cite{kong2018analysis} carry out a series of experiments to explore the relationship between students' academic performance and their graduation whereabouts.
Uosaki et al \cite{uosaki2018seamless} develop a career support system to help the international student find a job in Japan through their log record and eBook reading.
Liu et al \cite{liu2017hierarchical} design a job recommendation service framework for university students. According to a student profiling based re-ranking rule, users are recommended a list of potential jobs.
Soumya et al \cite{soumya2017improve} make a framework to identify student's eligibility for a specific job by calculating the domain competencies and job competency score.

\subsection{Dropout in Recurrent Neural Network}

Dropout is a mechanism that stops a part of neurons to improve generalization performance \cite{srivastava2014dropout}.
Zaremba et al \cite{zaremba2014recurrent} apply dropout technology on RNN with 0 memory loss through only applying the dropout operator on the non-recurrent connection.
Moon et al \cite{moon2015rnndrop} propose an effective solution to better preserve memory when applying dropout through generating a mask on input sequence and moving it at every time step.
Gal et al \cite{gal2016theoretically} propose a Bayesian interpretation-based RNN dropout variant method. They generate a dropout mask according to the theory of the Bayesian posterior and keep the same mask for each time step in the sequence.
Zhu et al \cite{zhu2016co} propose a dropout method for multilayer LSTM. To keep information stored in memory, they only allow the dropout mechanism to flow along with layers and prohibit it to flow along with the timeline.
Billa Jayadev \cite{billa2018dropout} test the dropout mechanism on LSTM algorithm based on a speech recognition system and the results with two datasets are improved about 24.64\% and 13.75\%, respectively.

\section{Problem Statement}
In this section, we will introduce some notations and then formally define the problem in this work. In a university, let ${\boldsymbol{\mathcal{M}}} = \{1,2,...,\textit{M}\}$ denote the set of majors and the set of students in every major is defined as $\boldsymbol{\mathcal{Q}}$ = \{$\mathcal{N}_1$,$\mathcal{N}_2$,...,$\mathcal{N}_M$\}. For student $i$ in major $m$, we define the \textbf {academic vector} as $\mathbf{a}_{i}^{m} \in \mathbb{R}^{n}$ that will be introduced in the following section. The \textbf {feature vector} and \textbf{final employment status}  are denoted as $\mathbf{d}_i^m \in \mathbb{R}^{p} $ and $y_i^m \in \{0,1\}$. Let $\mathbf{D}^m = [\mathbf{d}_1^m, \mathbf{d}_2^m,...\mathbf{d^m_{|\mathcal{N}_M|}}] \in \mathbb{R}^{|\mathcal{N}_M| \times p}$, $\mathbf{A}^{m} = [\mathbf{a}_1^{m}, \mathbf{a}_2^{m},...\mathbf{a}^{m}_{|\mathcal{N}_M|}] \in \mathbb{R}^{|\mathcal{N}_M| \times n}$ and $\mathbf{y}^m = [\mathbf{y}_1^m, \mathbf{y}_2^m,...\mathbf{y}^m_{|\mathcal{N}_M|}] \in \mathbb{R}^{|\mathcal{N}_M|}$ represent the feature matrix, the academic performance matrix and the employment status vector. The details of features used in this research are described in the following section.

\textbf{Employment Status Prediction Problem:}
given the feature vector $\mathbf{d}^m_i$ and the corresponding academic performance vector $\mathbf{a}^{m}_i$, then we predict the final employment status $\mathbf{y}^m_i$.

\begin{figure*}[t]
	\centering
    \includegraphics[width=1.9\columnwidth]{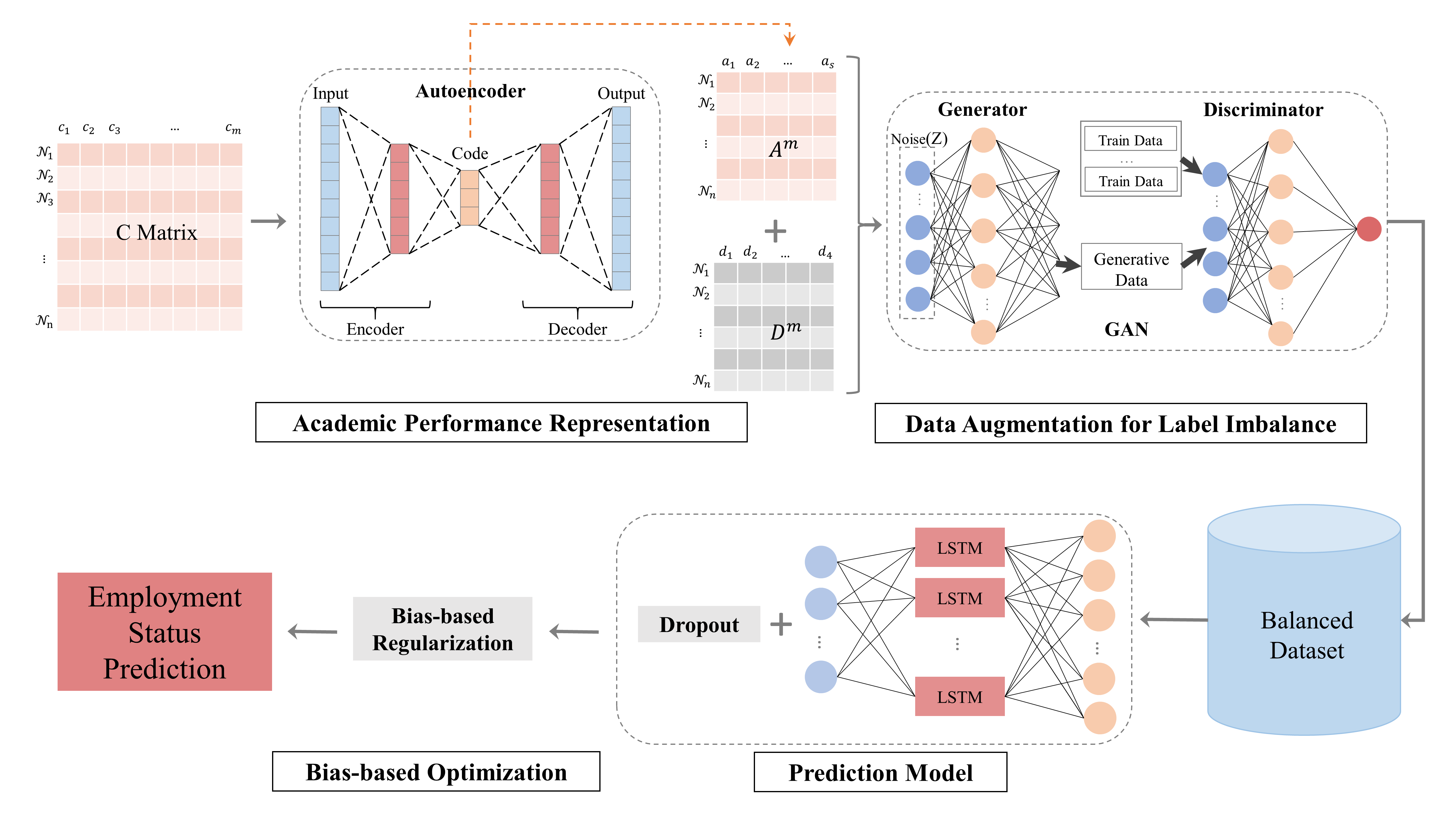}
	\caption{The illustration of MAYA.}
	\label{MAYA}
\end{figure*}

\section{Bias Analysis}
\subsection{Dataset}
The dataset used in this experiment includes 2,133 students from a Chinese university. They all enrolled in 2013 and graduated in 2017. They are from 64 different majors involved in 13 colleges. This dataset consists of three types of information, which are described as follows:
\subsubsection{Demographic Data}
Students are required to submit personal information at the time of admission, like hometown, gender, and nation. For the privacy concern, the students are already pseudonymous in the raw data. The demographic data includes 2,133 records.
\subsubsection{Academic Performance Data}
Students' academic performance data contains scores and credits of courses. There are in total 195,234 academic records.
\subsubsection{Employment Data}
When finding a job, students need to sign tripartite agreements to guarantee their legal rights, thus universities own records of students' employment status information about related companies and government agencies. This dataset consists of 2,133 records.

\subsection{Bias in Employment}
In this subsection, we analyze the bias in employment from two levels: major-level and college-level. We only show the results of the major-level and leave these of school-level in the Supplemental Material. We check the bias from four aspects: \textbf{gender} \cite{ford2018gender}, \textbf{nation} (the minority or the majority) \cite{al-ubaydli2019how}, \textbf{administrative level of hometown} (city or county) \cite{liang2018home}, and \textbf{enroll status} (whether passing the college entrance examination at one time). Chi-square test is used here to examine the impact of these features on 64 different majors involved in our dataset. Bias analysis of employment status is shown in Figure \ref{bias-major} . From the figure, the majors with bias in recruitment are shown as follows:
\begin{itemize}
\item \textbf{Gender}: English, Applied Psychology, Electronic Information Science and Technology.
\item \textbf{Administrative Level of Hometown}: Physical Education.
\item \textbf{Enroll Status}: Preschool education, English, Electronic Information Science and Technology, Food Science and Engineering.
\item \textbf{Nation}: Information and Computing Science, Computer Science and Technology.
\end{itemize}
These observations suggest that employment bias do exists in some majors and the existence of bias does affect graduates' employment. Note that we also analyze the bias in employment choice and results are provided in the Supplemental Material.

\section{ The Proposed MAYA Prediction Framework}
In this section, we provide a detailed description of the proposed framework, MAYA. Figure \ref{MAYA} shows an illustration of the MAYA framework. The framework has four components including representation learning of academic performance, data augmentation for label imbalance, prediction model, and bias-based optimization. Next we detail each component.

\subsection{Academic Performance Representation}
When taking academic performance as features, the heterogeneity of curriculum is always a challenge, due to the difference in students who take courses for each semester. In this work, we propose a $\boldsymbol{C}$ matrix and based on $\boldsymbol{C}$, we use an auto-encoder to get the embedding representation to tackle the heterogeneity.
\subsubsection{$\boldsymbol{C}$ Matrix}
To solve the problem caused by the heterogeneity, we create the matrix $\boldsymbol{C}_s$ $\in \mathbb{R}^{n_s \times m_s}$ where $n_s$ and $m_s$  represent the number of students and the number of courses, respectively. In our dataset, $s=1,2,...6$ since students have valid grades of 6 semesters except for the two-semester graduation project and social practice. $c_{ij}$ is the grade of student $i$ on course $j$. If a student does not attend a particular course, the corresponding element remains 0. The size of this matrix is different for each semester as shown in the following matrix. For example, there are 300 students and 500 courses in the first semester, then the size of $\boldsymbol{C_1}$ matrix is $300\times500$.

$$
\left\{
\begin{matrix}

 c_{11}      & c_{12}      & \cdots & c_{1n}\\
 c_{21}      & c_{22}      & \cdots & c_{2n}\\
  \vdots      & \vdots      & \ddots & \vdots \\
 c_{m1}      & c_{m2}      & \cdots &  c_{mn}\\

\end{matrix}
\right\}
$$

\subsubsection{Representation Learning}
$\boldsymbol{C}$ matrix is quite sparse, thus we use the autoencoder to get the embedding representation that is the academic performance matrix $\mathbf{A}$. The hidden layers of the autoencoder are divided into two parts: the encoder part and the decoder part. The layers consistently encode and decode the input data. The input of the $i$th layer is considered as the output of $(i-1)$th layer. The hidden layers can automatically capture the characteristics of input data and keep them unchanged. To capture the temporality among semesters, we use autoencoder to embed the matrix of each semester, respectively. In each hidden layer, we adopt the following nonlinear transformation function:
\begin{equation}
\begin{aligned}
&\boldsymbol{h_{(2)}} = f(\boldsymbol{W}_{(2)}\boldsymbol{h}_{(1)} + \boldsymbol{b}_{(2)}) \\
&\boldsymbol{h_{(3)}} = f(\boldsymbol{W}_{(3)}\boldsymbol{h}_{(2)} + \boldsymbol{b}_{(3)}) \\
&\boldsymbol{h_{(i)}} = f(\boldsymbol{W}_{(i)}\boldsymbol{h}_{(i-1)} + \boldsymbol{b}_{(i)}),i=1,2,...k
\end{aligned}
\end{equation}
where $f$ is the activation function and $\boldsymbol{W}_{(i)}$, $\boldsymbol{b}_{(i)}$ are the transformation matrix and the bias vector. We use $\boldsymbol{C}$ as the input and minimize the reconstruction error between the output and the original input. Then, we take the output of the encoder as the academic performance matrix $\mathbf{A}$.

\subsection{Data Augmentation for Label Imbalance}
In general, the number of students who fail to land a job is smaller, leading to a label imbalance problem. Thus, we employ generative adversarial networks (GAN) \cite{goodfellow2014generative} to augment data in order to improve the generalization performance. GAN consist of two components: a generator $G$ and a discriminator $D$ that compete in a two-player mini-max game on $V(D,G)$:
\begin{equation}
\begin{aligned}
&\mathop{\min}_{G}  \mathop{\max}_{D} V(D,G) = \mathbb{E}_{x \sim p_{data}} [logD(x)] + \\ &\mathbb{E}_{x \sim p_{z}(z)} [log(1-D(G(z))]
\end{aligned}
\end{equation}
The generator $G$ shown in Figure \ref{MAYA} takes a random vector from a uniform distribution as input. It outputs a vector including all features of the corresponding class (i.e., the students failed in job hunting). Then, the generated data and the real data are entered into discriminator $D$ for classification. Through repeated training, the $D$ can not identify the generated data from the real data. Then we use $G$ to generate data of students failed in job hunting until the two categories are balanced. In other words, we aim to implicitly learn the distribution of data of students failed in job hunting, to further generate new samples.

\subsection{Prediction Model}

\begin{figure}
	\centering
	\includegraphics[width= .95\columnwidth]{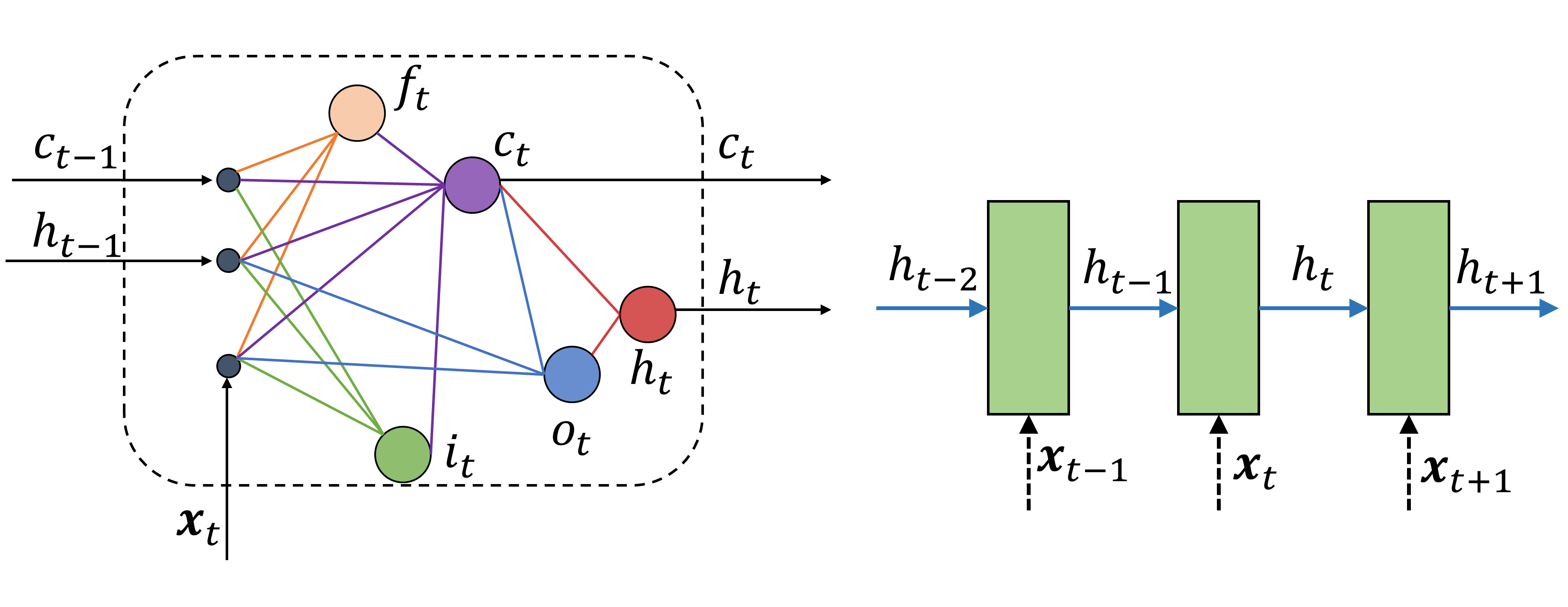}
	\caption{The diagram of LSTM.}
	\label{LSTM}
\end{figure}

We utilize LSTM to capture the sequentiality between semesters for prediction. LSTM is an RNN architecture that uses a vector of cells $\boldsymbol{c}_t \in \mathbb{R}^{n}$ with several element-wise multiplication gates to manage information. Generally, dropout aims to combine many 'thinned' networks to improve the performance of prediction. Some neurons are randomly dropped during the training stage in order to force the remaining sub-network to compensate. Then all the neurons are used for predictions during the testing stage.

In this work, we design a temporal dropout structure to improve the generalization performance. Previous researches show that it's not expected to erase information from a unit who remembers events that occurred many timestamps back in the past \cite{pham2014dropout,zhu2016co}. However, using complex models on a relatively simple dataset can easily cause overfitting. Moreover, the time span is not very long in our problem. Thus we allow the information of dropout in LSTM to flow along the time dimension. We utilize the classical LSTM framework shown as follows:
\begin{equation}
%\label{LSTME}
\begin{aligned}
&\boldsymbol{i}_t = \sigma(\boldsymbol{W}_{xi}\boldsymbol{x}_t + \boldsymbol{W}_{hi}\boldsymbol{h}_{t-1} + \boldsymbol{W}_{ci} \boldsymbol{c}_{t-1} + \boldsymbol{b}_i)  \\
&\boldsymbol{f}_t = \sigma(\boldsymbol{W}_{xf}\boldsymbol{x}_t + \boldsymbol{W}_{hf}\boldsymbol{h}_{t-1} + \boldsymbol{W}_{cf} \boldsymbol{c}_{t-1} + \boldsymbol{b}_f)  \\
&\boldsymbol{c}_t = (\boldsymbol{f}_t \odot \boldsymbol{c}_{t-1} + \boldsymbol{i}_t \odot tanh (\boldsymbol{W}_{xc} + \boldsymbol{W}_{hc}\boldsymbol{h}_{t-1} +\boldsymbol{b}_c ))  \\
&\boldsymbol{o}_t = \sigma(\boldsymbol{W}_{xo}\boldsymbol{x}_t + \boldsymbol{W}_{ho}\boldsymbol{h}_{t-1} + \boldsymbol{W}_{co} \boldsymbol{c}_{t} + \boldsymbol{b}_o) \\
&\boldsymbol{h}_t = \boldsymbol{o}_t \odot tanh(\boldsymbol{c}_t)
\end{aligned}
\end{equation}
where $\sigma (x)$ is the sigmoid function defined as $\sigma (x)= \frac{1}{1+e^{-x}}$. $\boldsymbol{W}_{\alpha\beta}$ denotes the weight matrix between $\alpha$ and $\beta$ (e.g., $\boldsymbol{W}_{xi}$ is the weight matrix from input $\boldsymbol{x}_t$ to the input gate $\boldsymbol{i}_t$), $\boldsymbol{b}_\alpha$ is the bias term of $\alpha \in \boldsymbol{\{} i,f,c,o   \boldsymbol{\}}$.

Inspired by \cite{zhu2016co}, we design an LSTM variant that allows the information of dropout in LSTM to flow along the time dimension through designing the mask vector $\boldsymbol{m}$ to drop the gates. The structure is shown in Figure \ref{LSTM}, defined by the following equations:
\begin{equation}
\label{LSTME}
\begin{aligned}
&\boldsymbol{i}_t = \sigma(\boldsymbol{W}_{xi}\boldsymbol{x}_t + \boldsymbol{W}_{hi}\boldsymbol{h}_{t-1} + \boldsymbol{W}_{ci} \boldsymbol{c}_{t-1} + \boldsymbol{b}_i) \odot \boldsymbol{m}_i \\
&\boldsymbol{f}_t = \sigma(\boldsymbol{W}_{xf}\boldsymbol{x}_t + \boldsymbol{W}_{hf}\boldsymbol{h}_{t-1} + \boldsymbol{W}_{cf} \boldsymbol{c}_{t-1} + \boldsymbol{b}_f) \odot \boldsymbol{m}_f \\
&\boldsymbol{c}_t = (\boldsymbol{f}_t \odot \boldsymbol{c}_{t-1} + \boldsymbol{i}_t \odot tanh (\boldsymbol{W}_{xc} + \boldsymbol{W}_{hc}\boldsymbol{h}_{t-1} +\boldsymbol{b}_c ))\\&  \odot \boldsymbol{m}_c \\
&\boldsymbol{o}_t = \sigma(\boldsymbol{W}_{xo}\boldsymbol{x}_t + \boldsymbol{W}_{ho}\boldsymbol{h}_{t-1} + \boldsymbol{W}_{co} \boldsymbol{c}_{t} + \boldsymbol{b}_o) \odot \boldsymbol{m}_o \\
&\boldsymbol{h}_t = \boldsymbol{o}_t \odot tanh(\boldsymbol{c}_t) \odot \boldsymbol{m}_h
\end{aligned}
\end{equation}
where $\odot$ represents element-wise product and $\boldsymbol{m}_f$, $\boldsymbol{m}_c$, $\boldsymbol{m}_o$ and $\boldsymbol{m}_h$ are dropout binary mask vectors, with an element value of 0 indicating that dropout happens, for input gates, forget gates, cells, output gates and output gates, respectively.

In our case, since we use single-layer LSTM, we only need to consider error back-propagation in the same network layer and the errors to output responses $\boldsymbol{h}_t$ are:
 \begin{equation}
\begin{aligned}
 &\boldsymbol{\epsilon}_{h}^{t} = \boldsymbol{\epsilon}_{h+1}^{t}\odot {m}_h  \\
 \end{aligned}
\end{equation}
where $\boldsymbol{\epsilon}_{h+1}$ represents the back-propagation error vectors from the next time in the same network layer.

Based on the Eq. \ref{LSTME}, we get the errors from  $\boldsymbol{h}_t$ to $\boldsymbol{o}_t$ which represents the errors from next time with dropout involved:
\begin{equation}
\begin{aligned}
 &\boldsymbol{\epsilon}_{o}^{t} = (\boldsymbol{\epsilon}_{h}^{t}\odot \frac{\partial \boldsymbol{h}_t}{\partial \boldsymbol{o}_t})\odot \boldsymbol{m}_o =  \boldsymbol{\epsilon}_{h}^{t} \odot tanh(\boldsymbol{c}_t) \odot \boldsymbol{m}_o \\
 \end{aligned}
\end{equation}
Using the same approach, we can get the back-propagation errors of other gates and the details are provided in the Supplemental Material.

\subsection{Bias-based Optimization}
\subsubsection{Modeling Employment Bias}
As mentioned above, bias is varied by majors in employment. This finding motivates us to eliminate the influence of bias in various majors. Therefore, we propose a smoothed regularization for the steady change of weight, defined as follow:
\begin{equation}
\begin{aligned}
 &\boldsymbol{\Omega}_{M} = \frac{1}{2}\sum_{m=1}^{M}\sum_{n>m}^M||\boldsymbol{W}\cdot(\boldsymbol{u}_m - \boldsymbol{u}_n)||_F^2 \\
 \end{aligned}
 \label{zxec}
\end{equation}
where $\boldsymbol{W}$ is the weight matrix of LSTM mentioned above. $||\cdot||_F^2$ denotes the Frobenius norm. $\mathbf{u}_m \in \mathbb{R}^{p}$ indicates the importance of the $p$ tested aspects in prediction for students in major $m$. It is represented by the $p$-value of chi-square test calculated in the Section 4 through a transformation function for emphasizing the importance of biases. In other words, the lower the $p$ value, the greater the weight of the bias. The transformation function is defined as follows:
\begin{equation}
\begin{aligned}
 &\boldsymbol{f(u)} = \frac{\boldsymbol{e}^{1-u}-\boldsymbol{e}^{1+u}}{\boldsymbol{e}^{1-u}+\boldsymbol{e}^{1+u}} \\
 \end{aligned}
\end{equation}

\subsubsection{Optimization}

\begin{figure}
	\centering
	\includegraphics[width=.95\columnwidth]{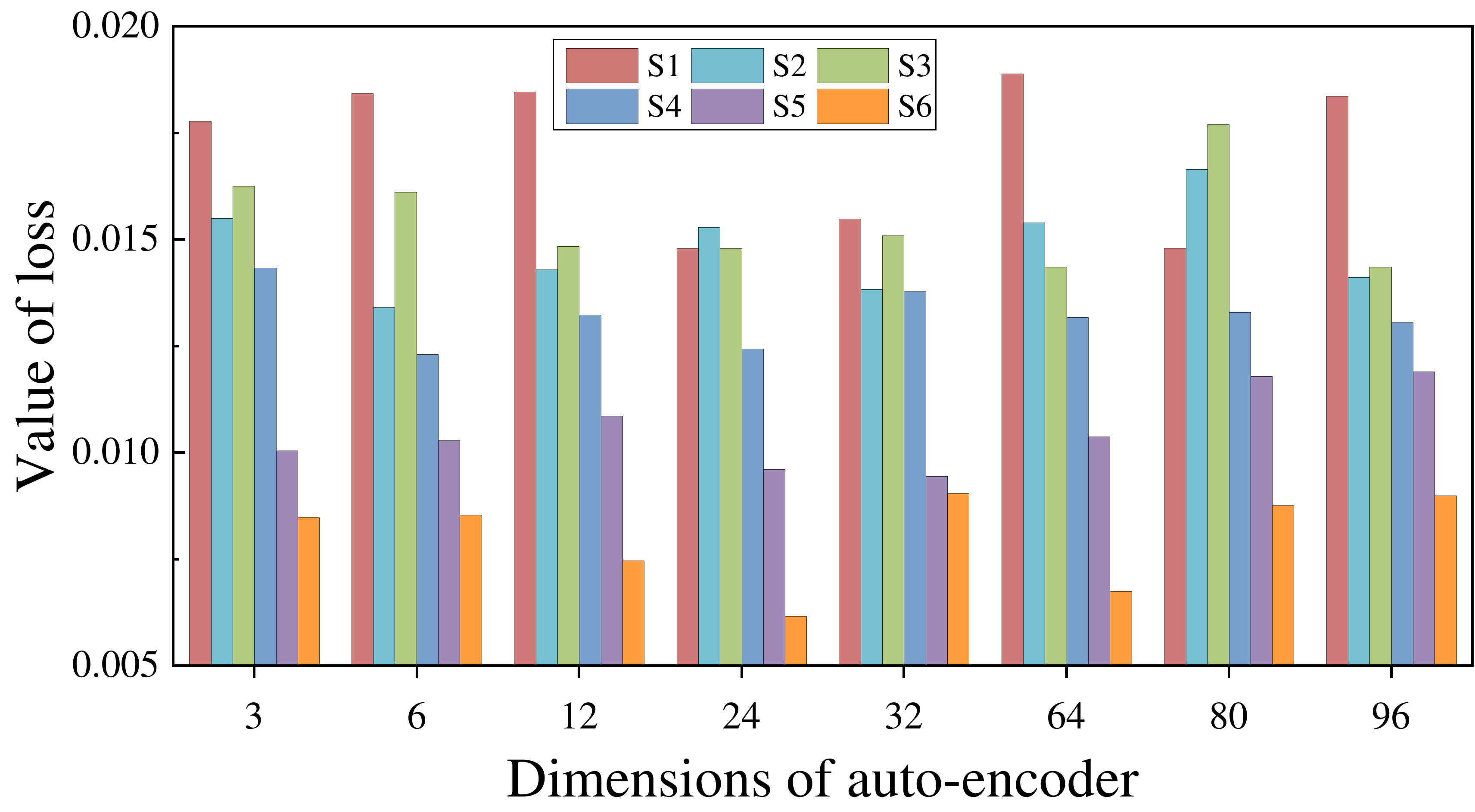}
	\caption{The results of representation learning.}
	\label{autoencoder}
\end{figure}

Based on the discussion above, we formulate the whole loss function of our MAYA prediction framework as follows:

\begin{equation}
\begin{aligned}
 \boldsymbol{\mathcal{L}} &= \frac{1}{2}\sum_{m=1}^{M}(\boldsymbol{W}\boldsymbol{x}_i^m - \boldsymbol{y}_i^m)^2 + \boldsymbol{\Omega}_{M}
 \end{aligned}
 \label{bias-opt}
\end{equation}

Its corresponding gradient is shown as follows:
\begin{equation}
\begin{aligned}
\nabla\boldsymbol{\mathcal{L}}(\boldsymbol{W}) = \sum_{m=1}^{M}\boldsymbol{W}\boldsymbol{x}_i^m (\boldsymbol{x}_i^{m})^{\mathsf{T}} - \boldsymbol{y}_i^m (\boldsymbol{x}_i^m)^{\mathsf{T}} + \\ \sum_{m=1}^{M}\sum_{n>m}^N \boldsymbol{W}(\boldsymbol{u}_m - \boldsymbol{u}_n)(\boldsymbol{u}_m - \boldsymbol{u}_n)^{\mathsf{T}}
 \end{aligned}
\end{equation}

\begin{figure*}
\centering
\includegraphics[width=0.63\columnwidth]{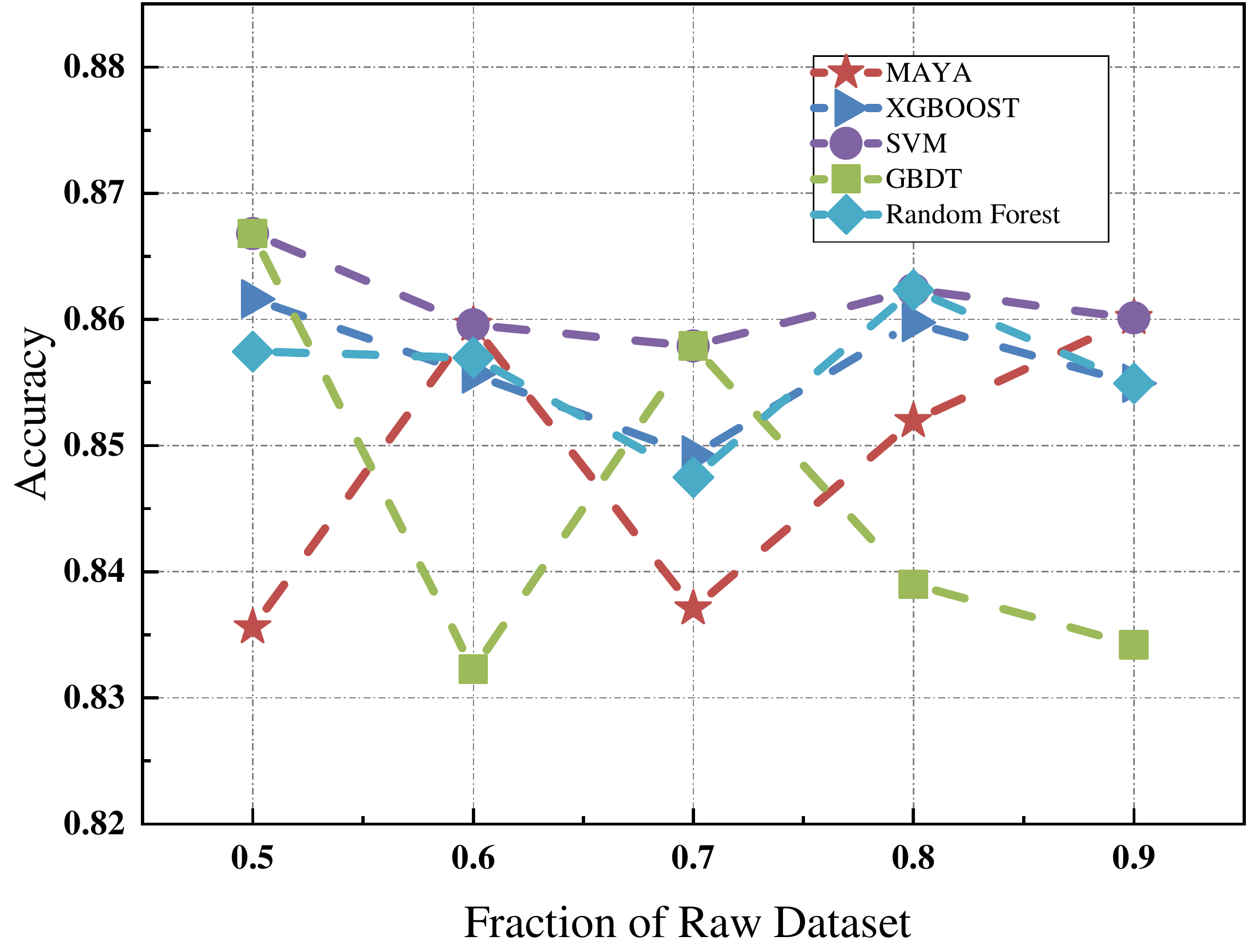}
\includegraphics[width=0.63\columnwidth]{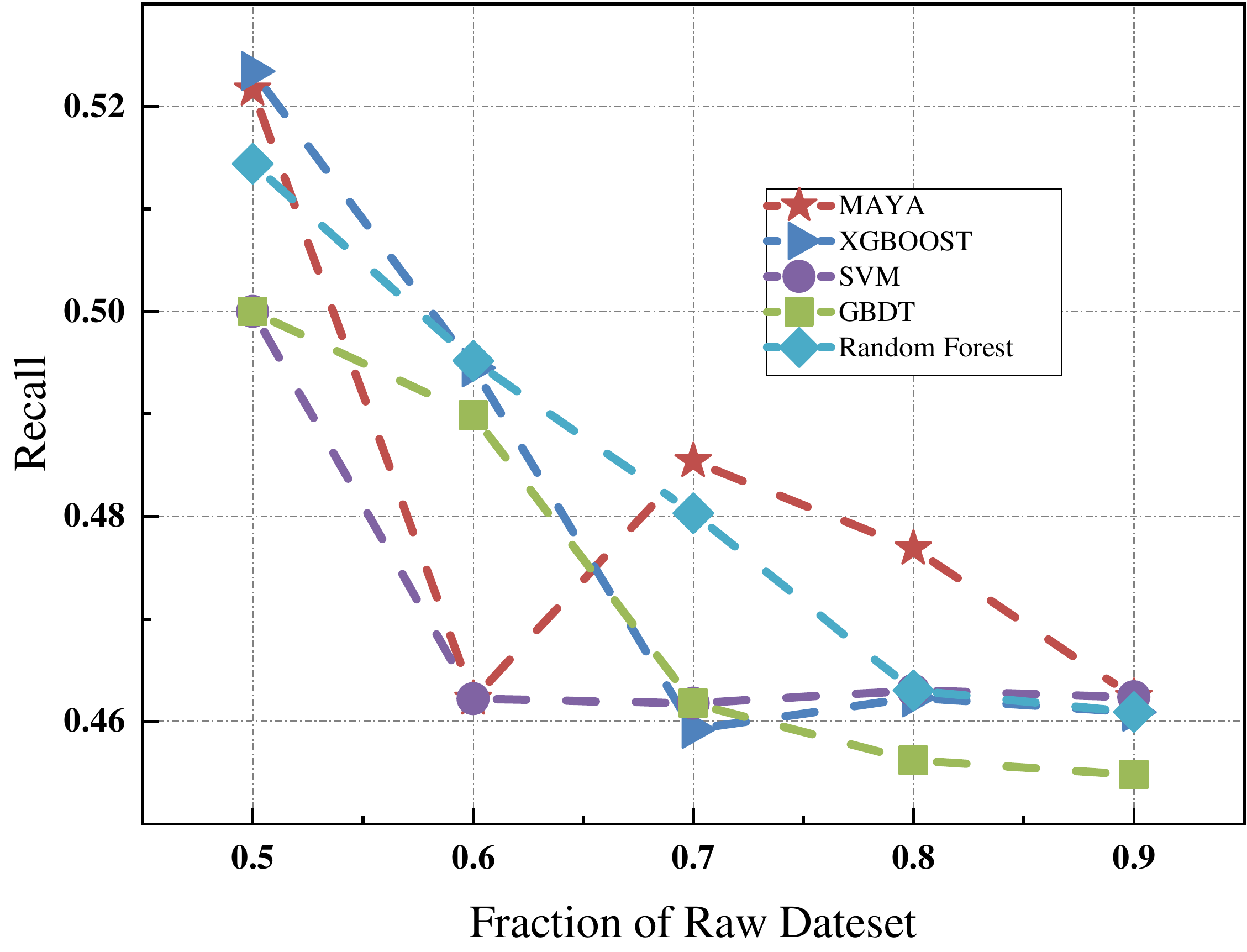}
\includegraphics[width=0.63\columnwidth]{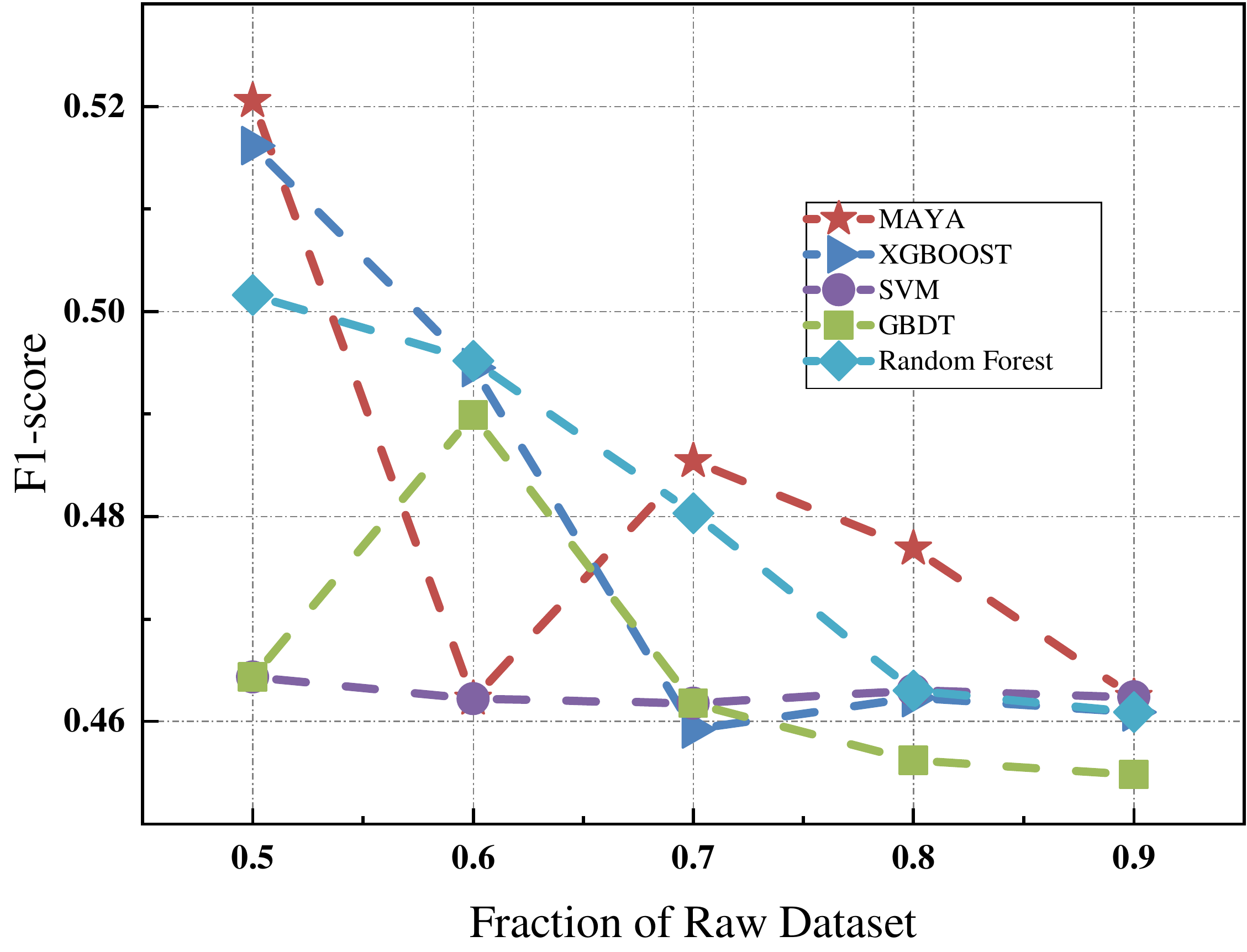}
\caption{Prediction performance on raw training dataset $a$.}
\label{raw}
\end{figure*}

\begin{figure*}
\centering
\includegraphics[width=0.63\columnwidth]{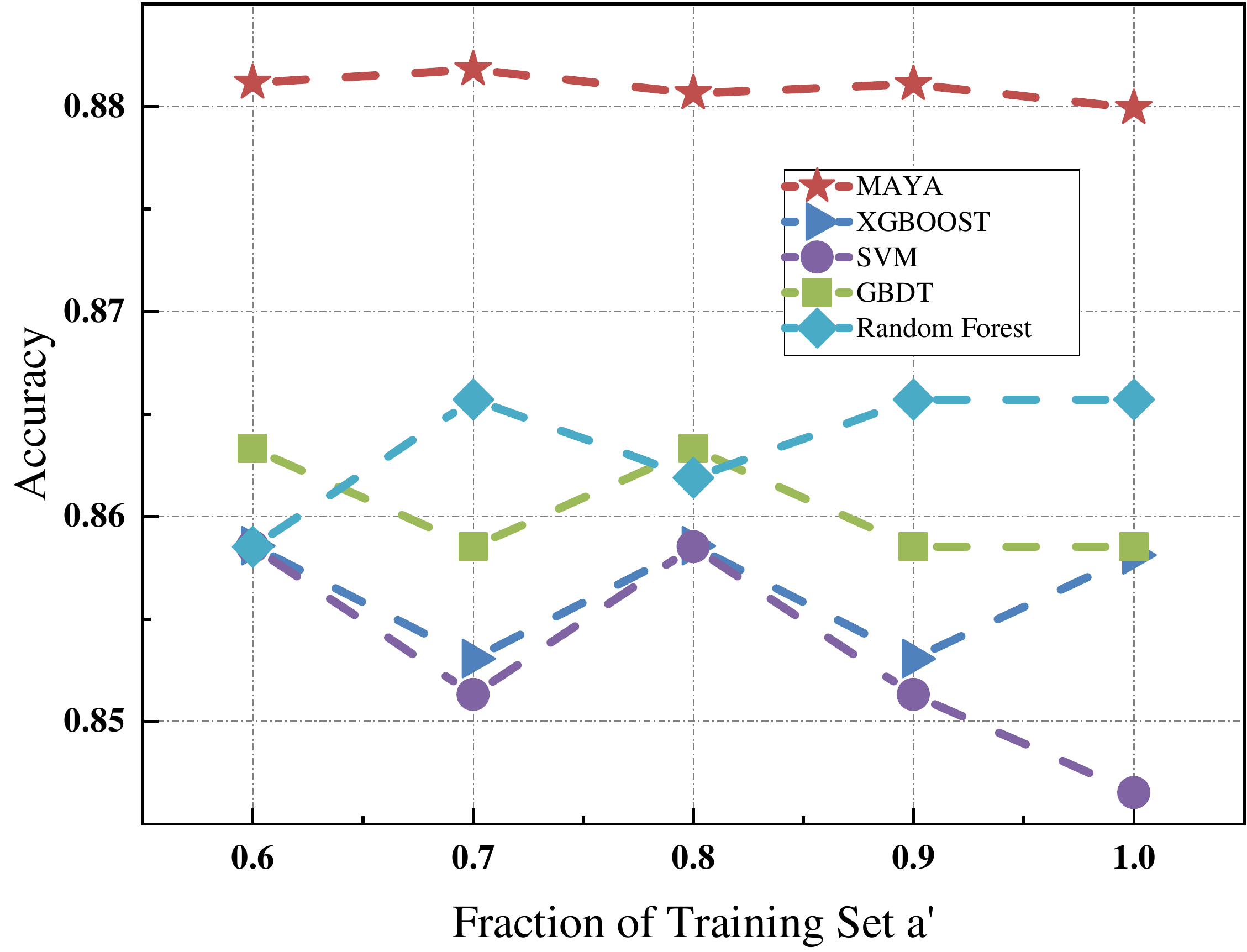}
\includegraphics[width=0.63\columnwidth]{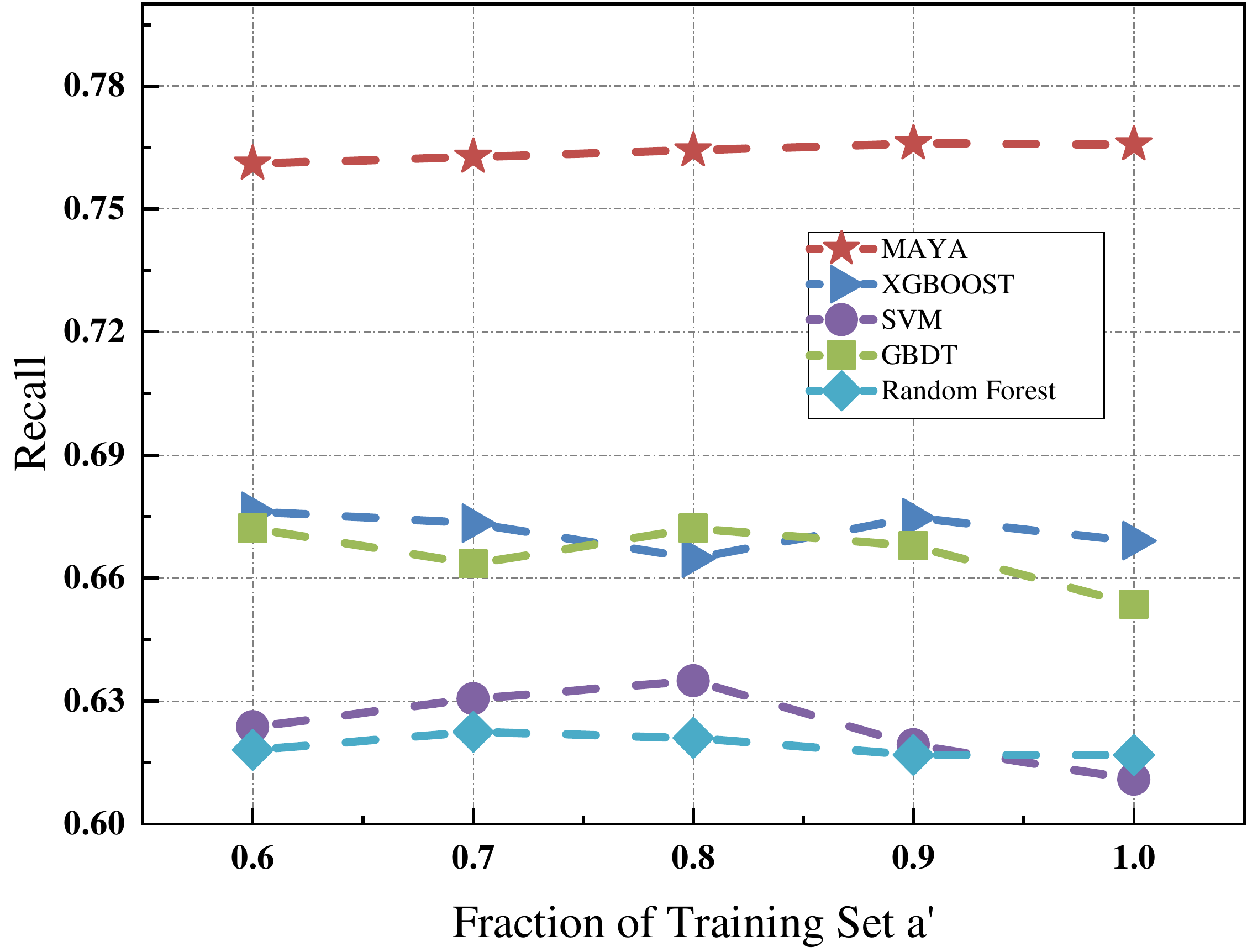}
\includegraphics[width=0.63\columnwidth]{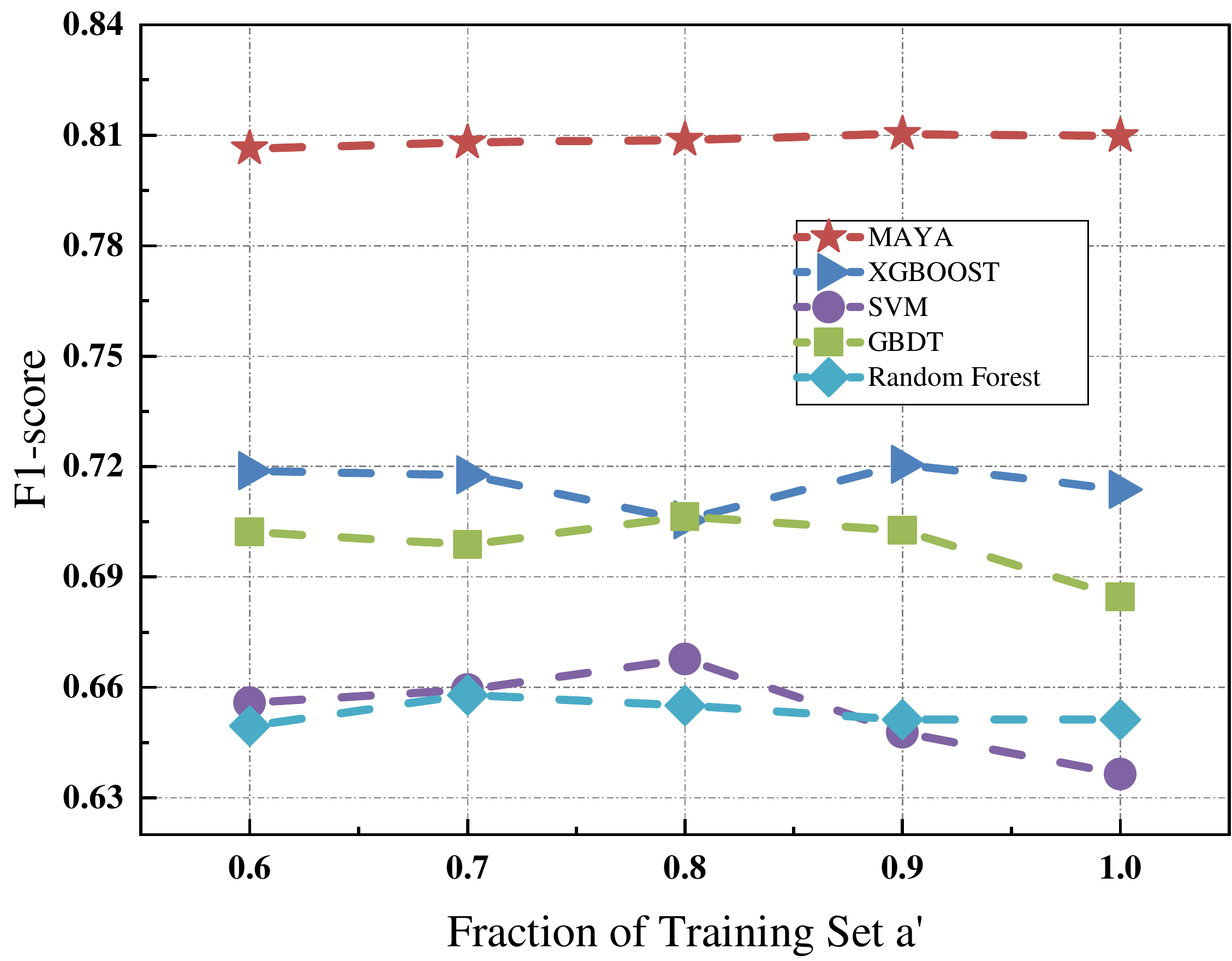}
\caption{Prediction performance on balanced training dataset $a'$.}
\label{bal}
\end{figure*}

\section{Experiment}
In this section, we would present the experimental results in detail to demonstrate the effectiveness of our proposed MAYA. We first introduce the experimental settings, then present comparison results and finally investigate important parameters of the proposed framework.

\subsection{Experimental Settings}
To deal with the heterogeneity of courses enrolled by students each semester, we design a $\boldsymbol{C}$ matrix to denote students' academic performance. The four-year university life involves 8 semesters. Campus recruitment takes place densely at the beginning of the last year. Hence, only the academic performance of the previous three years (or six semesters $S1$ to $S6$) would affect students' employment. Autoencoder is applied to embed academic performance data to overcome the heterogeneity of course selection. We test the different dimensions including 3, 6, 12, 24, 32, 64, 80, 96 and the performance is shown in Figure \ref{autoencoder}. The value of loss function fluctuates slightly. That is, even vectors with low dimensions can still effectively represent the academic performance of each student. Thus, we choose 3 as the dimension of representation for computational efficiency.

\subsection{Prediction Results}
 We predict employment status with features including academic performance, gender, nation, enroll status, hometown and their major. To verify the effectiveness of our MAYA framework, we design prediction experiments including two settings: comparison with LSTM-based MAYA's variants and comparison with representative baselines.

\subsubsection{Comparison with LSTM-based MAYA's Variants}

Table \ref{variant} displays the prediction performance of MAYA and its variants. We design a four-step experiment to test the performance with metrics, i.e., accuracy, recall and F1-score, to understand the results collectively.

\begin{table}
\begin{tabular}{lllllll}
\hline
Variants& Accuracy   & Recall & F1-score &   \\
\hline
LSTM+Raw Data& 0.862  & 0.500 & 0.463 &   \\
LSTM+GAN & 0.869  & 0.670 & 0.717 &  \\
\makecell[l]{LSTM+Dropout\\+GAN}& 0.876  & 0.712 & 0.761 & \\
\makecell[l]{LSTM+Dropout+\\GAN+New Loss}& 0.880  & 0.766 & 0.810 & \\
\hline
\end{tabular}
\caption{Prediction performance of MAYA variants.}
\label{variant}
\end{table}

In the first step, we use raw data to fit the LSTM algorithm. The imbalance label issue exists in raw data, leading to the occurrence of unexpected results on precision and recall. First, the algorithm could achieve a low loss value by ignoring the minority class and predicting all the samples into the majority class. It results in a recall with 0.5. Second, the precision is relatively low because no samples are predicted as a minority. In other words, for the minority class, the correct ratio of prediction results is 0.

In the second step, GAN is used to solve the label imbalance and the data generation process is shown as follows:
\begin{itemize}
\item First, raw data is divided into two categories: training set $\boldsymbol{a}$ and testing set $\boldsymbol{b}$ by stratified sampling.
\item Second, we use GAN on the training set $\boldsymbol{a}$ to generate samples of the minority class. Then in the new training set $\boldsymbol{a}'$, the number of students in the two classes is equal.
\end{itemize}

Then, we use the training set $\boldsymbol{a}'$ to fit the model and test it on the original testing set $\boldsymbol{b}$. The performance shown in Table \ref{variant} verifies its effectiveness.

In the third step, a new dropout mechanism of LSTM is employed to alleviate the overfitting problem caused by the relatively small experimental dataset. We add the dropout mechanism based on the experiment of the second step. In the final step, we add the bias-based regularization into the optimization loss based on the last step and the results suggest its importance.

\subsubsection{Comparison with Baseline Methods} In addition to the comparison with deep learning-based variants, we compare the MAYA framework with several popular algorithms shown as follows:
\begin{itemize}
\item \textbf{SVM} \cite{scholkopf2001learning}: SVM is a classic algorithm and is widely used in the field of data mining.
\item \textbf{Random Forest} \cite{breiman2001random}: is a classic ensemble algorithm that achieves good performance in various applications.
\item \textbf{GBDT} \cite{friedman2001greedy}:  GBDT is an additive regression model consisting of regression trees.
\item \textbf{XGBoost} \cite{chen2016xgboost}: XGBoost is a boosting-tree-based method and is widely used in various data mining scenarios with good performance.
\end{itemize}

Note that, we test the performance of these algorithms from two aspects. On one hand, we fit algorithms based on the raw training set $a$ and test them on testing set $b$. The results are shown in Figure \ref{raw}. It's shown that the prediction is not accurate and the fluctuation is quite large due to the label imbalance of raw data. To overcome this problem, we fit algorithms based on the balanced training set $a'$ and test them on $b$. The results are shown in Figure \ref{bal}. The performance is improved significantly.

\subsection{Parameter Sensitivity}
\subsubsection{Dropout Proportions}

\begin{figure}
	\centering
	\includegraphics[width=.95\columnwidth]{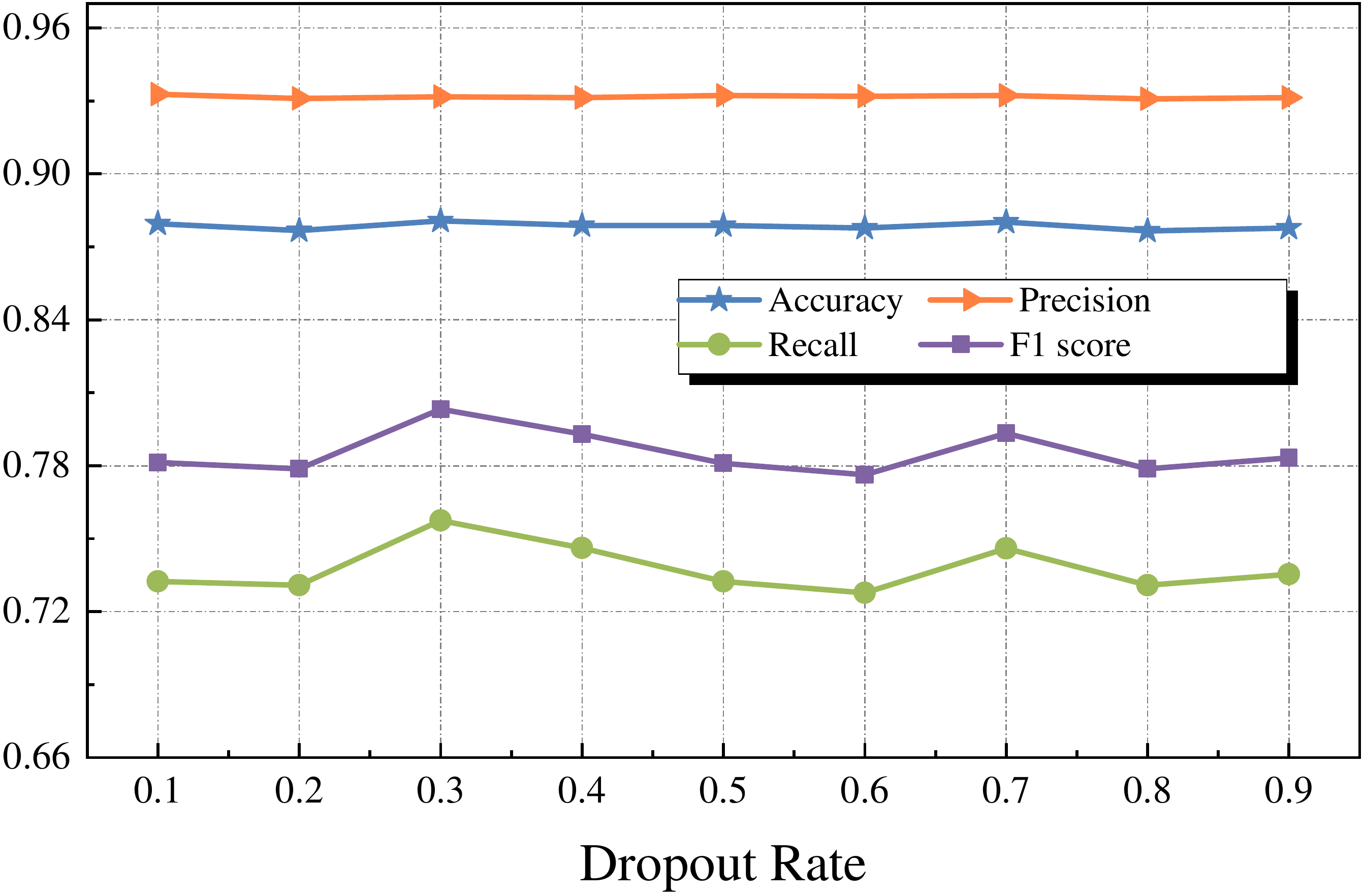}
	\caption{Performance of MAYA with different dropout proportions.}
	\label{dropout}
\end{figure}

As mentioned above, a dropout mechanism is utilized to improve the generalization performance and we test the sensitivity of MAYA framework on dropout proportions (Figure \ref{dropout}). The change of dropout proportions generates a slight impact and 0.3 is the best.

\subsubsection{Input Features}

It's of great significance to distinguish early the students who might encounter difficulties in employment. Then teachers can intervene at an early stage. Therefore, we conduct a test on the number of semesters involved in academic performance data. As shown in Table \ref{semester}, prediction performance grows slowly since the fourth semester.
In other words, we can predict the students with trouble in landing a job with high accuracy at the end of the second year.
We also design an experiment to test the effectiveness of demographic features and leave the results in Supplemental Material.

\begin{table}
  \centering
    \resizebox{.95\columnwidth}{!}{
    \begin{tabular}{lllll}
    \hline
    \multicolumn{1}{l}{Semester} & \multicolumn{1}{l}{Accuracy} & \multicolumn{1}{l}{Precision} & \multicolumn{1}{l}{Recall} & \multicolumn{1}{l}{F1} \\
    \hline
    1     & 0.73469 & 0.60848& 0.62668 & 0.61484 \\
    2     & 0.82287 & 0.71878 & 0.65398 & 0.67470 \\
    3     & 0.86666 & 0.92909 & 0.654762 & 0.69820 \\
    4     & 0.86758 & 0.92944 & 0.658824 & 0.70311 \\
    5     & 0.87631 & 0.93264 & 0.69898 & 0.74856 \\
    6     & 0.88073 & 0.93172 & 0.757463 & 0.80326 \\
    \hline
    \end{tabular}}%
\caption{Performance of MAYA with different numbers of semesters.}
\label{semester}%
\end{table}%

\subsubsection{Learning Rate}

\begin{figure}
	\centering
	\includegraphics[width=.95\columnwidth]{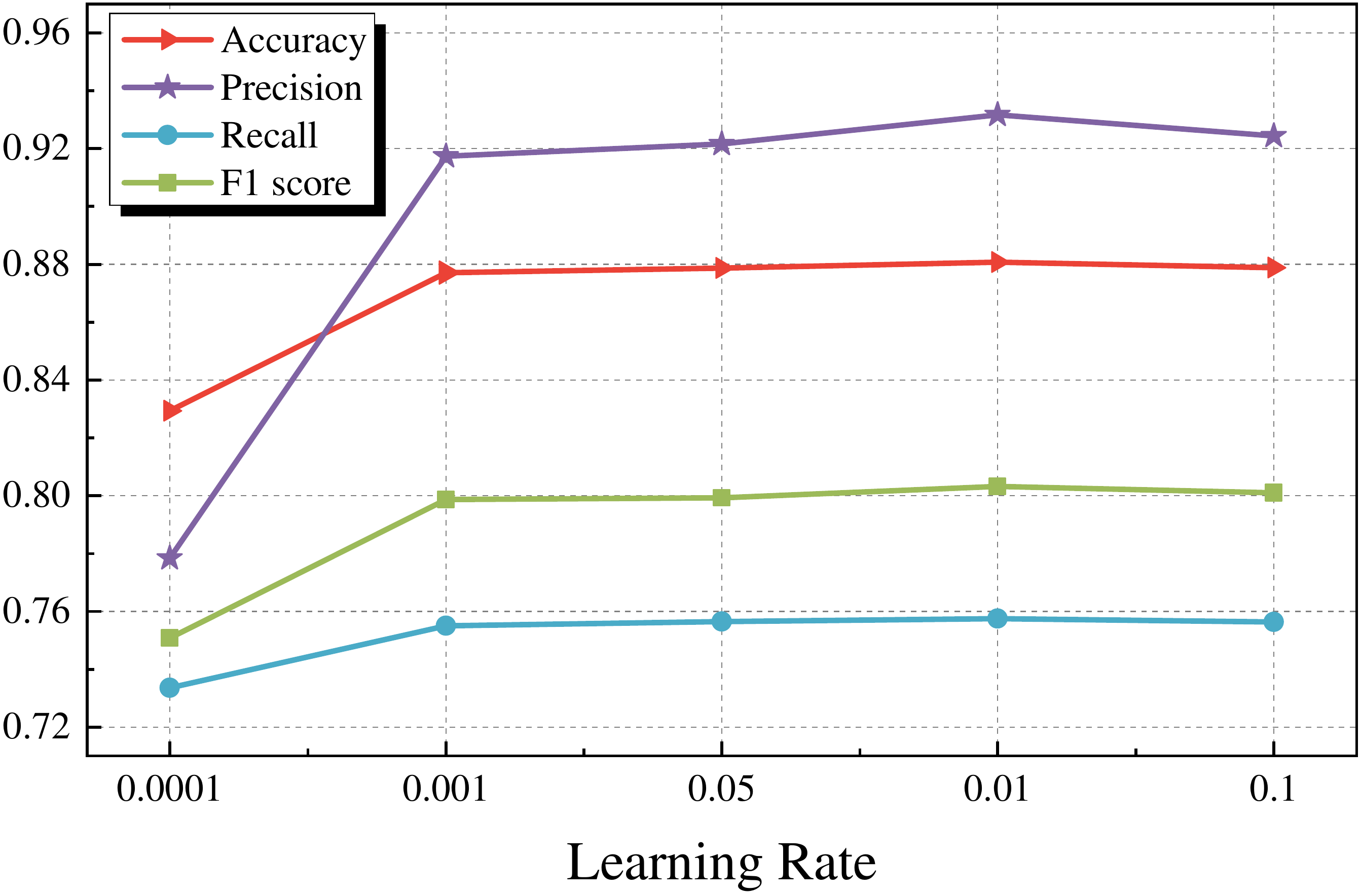}
	\caption{Performance of MAYA with different learning rates.}
	\label{learningrate}
\end{figure}

Learning rate that controls the update speed of the model is an important parameter in  the MAYA framework. In Figure \ref{learningrate}, we analyze the performance of various learning rates and find that the model can achieve the best prediction performance when the learning rate is set to 0.01.

\subsubsection{Bias-based Regularization}
We design an experiment to test the effectiveness of bias based regularization. We use Eq.\ref{zhengze} and Eq.\ref{bias-opt} as loss function separately and the prediction performance is shown in Table \ref{ceshilaosi1}. The bias-based regularization can improve the performance remarkably.

\begin{equation}
\begin{aligned}
 \boldsymbol{\mathcal{L}_1} &= \frac{1}{2}\sum_{m=1}^{M}(\boldsymbol{W}\boldsymbol{x}_i^m - \boldsymbol{y}_i^m)^2 + ||\boldsymbol{W}||_F^2 \\
  \end{aligned}
 \label{zhengze}
\end{equation}

\begin{table}
\center
\resizebox{.95\columnwidth}{!}{
\begin{tabular}{lllllll}
\hline
Optimization Function & Accuracy  & Recall & F1-score &   \\
\hline
Eq.\ref{zhengze}& 0.873  & 0.689& 0.738 &   \\
Eq.\ref{bias-opt} & 0.880  & 0.766 & 0.810 & \\
\hline
\end{tabular}}
\caption{Performance of Different Optimization.}
\label{ceshilaosi1}
\end{table}

\section{Conclusion and Discussion}

In this paper, we analyze a large-scale educational data for predicting graduates' employment status. For the reason that bias cannot be ignored in employment, we analyze the employment bias of different majors first of all and verify the existence of employment bias. Then based on such bias, MAYA, a prediction framework, is proposed in this paper to predict graduates' employment status. We incorporate the autoencoder to ease the data-sparse issue and deal with the imbalance label data using GAN. LSTM is combined with dropout and a bias-based regularization to overcome the over-fitting problem and capture the impact of biases.
Our extensive experiments based on an education dataset demonstrate the proposed framework can improve the prediction performance significantly and MAYA outperforms other baselines like LSTM and XGBoost significantly.

There are multiple directions for future work. Firstly, we plan to expand our dataset and explore this issue from more aspects. Secondly, we would conduct data acquisition from various companies and further study this issue from the company's perspective. Last but not least, we also intend to integrate MAYA framework into the modern educational management system and apply it to detect the employment status of graduating students.

\bibliography{6392-aaai}
\bibliographystyle{aaai}

\end{document}